\documentclass[journal]{IEEEtran}
\pdfoutput=1
\usepackage{amsmath,amsfonts}
\usepackage{algorithmic}
\usepackage{algorithm}
\usepackage{array}
\usepackage[caption=false,font=normalsize,labelfont=sf,textfont=sf]{subfig}
\usepackage{textcomp}
\usepackage{stfloats}
\usepackage{url}
\usepackage{verbatim}
\usepackage{graphicx}
\usepackage{cite}
\usepackage{tabularx}
\usepackage{booktabs} 
\usepackage{multirow}
\usepackage[hidelinks]{hyperref}

\begin{document}

\title{EGP3D: Edge-guided Geometric Preserving 3D Point Cloud Super-resolution for RGB-D camera}

\author{Zheng Fang\textsuperscript{\footnotesize 1}\IEEEauthorrefmark{2}, Ke Ye\textsuperscript{\footnotesize 2}\IEEEauthorrefmark{2}, Yaofang Liu\textsuperscript{\footnotesize 3}, Gongzhe Li\textsuperscript{\footnotesize 1}, Jialong Li\textsuperscript{\footnotesize 4}, Ruxin Wang\textsuperscript{\footnotesize 1}, Yuchen Zhang\textsuperscript{\footnotesize 1}, Xiangyang Ji\textsuperscript{\footnotesize 5} and Qilin Sun\textsuperscript{\footnotesize 1}\IEEEauthorrefmark{1}

\thanks{\IEEEauthorrefmark{2} Co-first authors.}
\thanks{\IEEEauthorrefmark{1} Corresponding author. }
\thanks{\textsuperscript{\footnotesize 1} School of Data Science, The
Chinese University of Hong Kong (Shenzhen), Shenzhen 518172, China}
\thanks{\textsuperscript{\footnotesize 2} The Hong Kong University of Science and
Technology (Guangzhou), Guangzhou 511458, China}
\thanks{\textsuperscript{\footnotesize 3} City University of Hong Kong, Tat Chee Avenue, Kowloon, Hong Kong SAR}
\thanks{\textsuperscript{\footnotesize 4} King Abdullah University of Science and Technology, Thuwal 23955, Saudi Arabia}
\thanks{\textsuperscript{\footnotesize 5} Tsinghua University, Beijing 100084, China}
}

% \markboth{IEEE TRANSACTIONS ON VISUALIZATION AND COMPUTER GRAPHICS,~Vol.~XX, No.~X, December~2024}%
\markboth{IEEE TRANSACTIONS ON XX,~Vol.~XX, No.~X, December~2024}{}%



\maketitle

\begin{figure*}[htpb]
\centering
\includegraphics[width=1\textwidth]{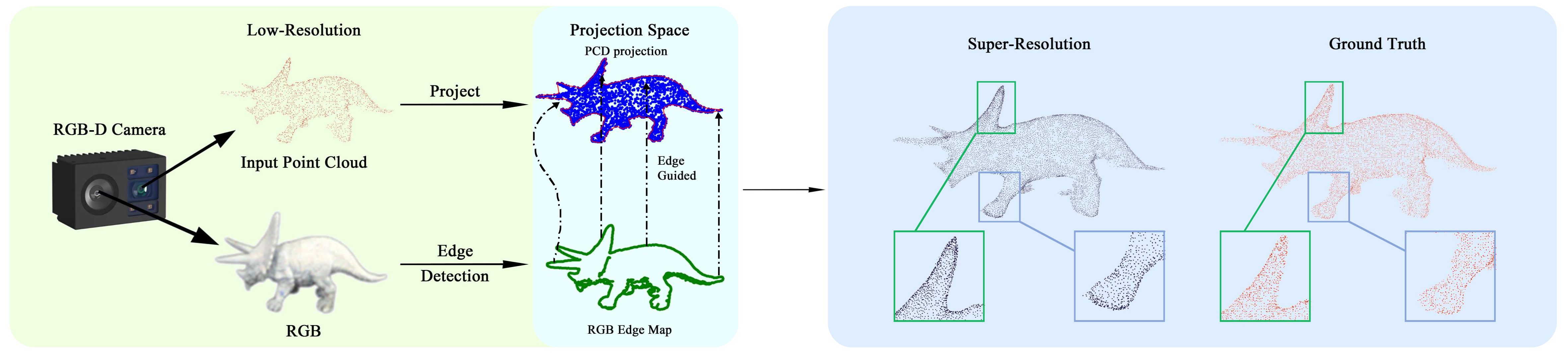}
\caption{Due to depth sensor resolution limitations, the captured point cloud has limited pixel counts (left). Our pipeline projects the point cloud into 2D space (center) and utilizes RGB image edge information for upsampling guidance. This enables us to reconstruct a high-fidelity, super-resolved point cloud (right) using a geometric preserving algorithm.}
\label{fig:teaser_figure}
\vspace*{-10pt}
\end{figure*}

\begin{abstract}
Point clouds or depth images captured by current RGB-D cameras often suffer from low resolution, rendering them insufficient for applications such as 3D reconstruction and robots. Existing point cloud super-resolution (PCSR) methods are either constrained by geometric artifacts or lack attention to edge details. To address these issues, we propose an edge-guided geometric-preserving 3D point cloud super-resolution (EGP3D) method tailored for RGB-D cameras. Our approach innovatively optimizes the point cloud with an edge constraint on a projected 2D space, thereby ensuring high-quality edge preservation in the 3D PCSR task. To tackle geometric optimization challenges in super-resolution point clouds, particularly preserving edge shapes and smoothness, we introduce a multi-faceted loss function that simultaneously optimizes the Chamfer distance, Hausdorff distance, and gradient smoothness. Existing datasets used for point cloud upsampling are predominantly synthetic and inadequately represent real-world scenarios, neglecting noise and stray light effects. To address the scarcity of realistic RGB-D data for PCSR tasks, we built a dataset that captures real-world noise and stray-light effects, offering a more accurate representation of authentic environments. Validated through simulations and real-world experiments, the proposed method exhibited superior performance in preserving edge clarity and geometric details.
\end{abstract}


\section{Introduction}

\IEEEPARstart{R}{GB-D} cameras have emerged as pivotal tools for capturing accurate point cloud data and revolutionizing domains, such as robotics and autonomous driving~\cite{sturm2012benchmark, premebida2019rgb, sun2018motion, wang2019self}. By fusing precise depth information with traditional RGB images, they enable advanced applications, ranging from object recognition to navigation and complex scene understanding. Nevertheless, RGB-D cameras often have limitations, producing low-resolution point clouds marred by unclear boundaries due to variations in light sensitivity across objects and backgrounds. Consequently, there is an urgent need to develop methods that can harness RGB-D cameras to generate dense uniform point clouds enriched with geometric intricacies.

In the realm of two-dimensional imaging, substantial progress has been made, particularly in edge detection, leading to a pertinent question: Can RGB images serve as a compass to navigate the challenges of point cloud resolution? Early endeavors, like guided depth map super-resolution (GDSR)~\cite{zhong2023guided}, attempted to steer the path, leveraging super-resolved color images to enhance depth map quality. However, the transition from depth maps to point clouds is fraught with variability, underscoring the need for direct point cloud super-resolution.

Point cloud upsampling, similar to the pursuit of super-resolution in point clouds, aims to transform sparse representations into dense geometrically coherent structures. Early efforts in this domain predominantly employed interpolative approaches \cite{yu2018pu, li2019pu, feng2022neural}, but these often failed to capture geometric subtleties, particularly along edges. While recent advancements have sought to address these limitations \cite{li2024learning, du2024arbitrary}, they have either focused on specific challenges such as filling holes or depending on computationally expensive 3D voxel networks. Consequently, a comprehensive solution ensures that the upsampled point clouds accurately reflect the shapes of real-world objects remain elusive.

Moreover, the reliance on synthetic datasets, such as PU-GAN \cite{li2019pu} and PU1K \cite{qian2021pu} limits the generalization of these methods. These datasets, which are devoid of real-world complexities such as noise, stray light effects, and data incompleteness, fail to adequately represent the challenges posed by RGB-D cameras in dynamic environments.

In this study, we embark on a novel trajectory, leveraging RGB data as a beacon to illuminate the path towards geometric refinement in point cloud super-resolution for RGB-D cameras.  Our proposed method is tailored specifically for RGB-D camera outputs, as shown in Figure \ref{fig:teaser_figure}. Initially, we employ the intrinsic and extrinsic parameters of the RGB-D camera to map the point clouds to their corresponding RGB images, establishing the foundation for our geometric optimization process. Subsequently, we extract the concave hull \cite{barber1996quickhull} of the point clouds, revealing the boundary of the projected points on the RGB image. Following this, we integrate the projected boundary of the point clouds with the object boundaries in the RGB images, transforming this challenge into a two-dimensional edge optimization problem. Ultimately, we design three loss functions specifically targeting this 2D problem: Chamfer distance loss \cite{remy2002medial}, Hausdorff distance loss \cite{rucklidge1996efficient}, and gradient smooth loss. These loss functions are seamlessly integrated into the existing upsampling model to achieve point cloud super-resolution.

Furthermore, we created a novel dataset captured directly by RGB-D cameras, encompassing a wide array of objects ranging from simple geometric shapes to complex figures, such as human portraits and dinosaur models. Each model was captured from multiple angles to ensure the comprehensiveness and complexity of the data. Compared with existing datasets, our dataset more accurately reflects the characteristics of point clouds in real-world scenarios, including noise, stray light, and other challenging factors. 

In summary, our contribution lies in three aspects: 
\begin{itemize}
    \item We propose the EGP3D method, enabling high-quality point cloud super-resolution with geometric preservation tailored for RGB-D cameras.\hfill
    \item We introduce an edge-guided module for point cloud super-resolution, accompanied by carefully designed loss functions. This innovative approach surpasses existing point cloud upsampling and depth-based super-resolution methods.\hfill
    \item We have captured a high-quality point cloud dataset using an RGB-D camera, specifically designed for point cloud super-resolution tasks. This dataset takes noise and stray light effects into consideration, enabling realistic generalization to real-world scenarios.\hfill
\end{itemize}

\section{Related Works}
\subsection{Learning-based Point Cloud Upsampling}
 Since the inception of PointNet \cite{Qi_2017_CVPR} and PointNet++ \cite{NIPS2017_d8bf84be}, the field of point cloud upsampling (PU) has been revolutionized. PU-Net \cite{yu2018pu} emerged as a pioneering work that introduced an upsampling network that laid the foundation for numerous subsequent innovations. Subsequently, many methods have been developed to improve the interpolation of points during upsampling \cite{yifan2019patch,qian2021deep,long2021pc2}. However, a significant shift occurred with the advent of MAFU \cite{9555219}, which redirected the field's attention towards arbitrary-scale upsampling. Subsequently, Feng et al. \cite{feng2022neural} and Dell'Eva et al. \cite{dell2022arbitrary} proposed unique arbitrary-scale upsampling methods.  
 
 Despite these advancements, a notable oversight persisted: most methods prioritized upsampling accuracy and rate, often neglecting the crucial aspect of the edge geometry. Recent studies have begun to address this geometric optimization, with Grad-PU \cite{he2023grad}, Du et al.'s voxel-based network \cite{du2024arbitrary}, RepKPU's kernel point convolution \cite{rong2024repkpu}, and Li et al.'s implicit fields \cite{li2024learning} contributing to this emerging trend. However, their focus was primarily on point placement rather than explicit edge geometry. Furthermore, the reliance on synthetic data limits the real-world applicability of these methods. In contrast, our EGP3D method distinguishes itself by emphasizing the edge geometry in the densification process, tackling the boundary optimization challenge that has been overlooked by previous approaches. 

\subsection{Guided Depth Map Super-Resolution} The concept of RGB-guided tasks first emerged within the realm of depth map super-resolution in the 3D vision domain.  Early works on GDSR, such as mutually guided image filtering \cite{8550683}, utilized multiple image information for filtering purposes. This approach was further refined by integrating color-guided internal and external regularizations \cite{8491336}, enhancing both resolution and quality through multi-direction dictionary learning and autoregressive modelling \cite{8861360}.

In recent years, the landscape of GDSR has shifted towards learning-based methods, and multimodal convolutional dictionary learning was introduced \cite{gao2022multi}, while geometric spatial aggregators were developed for continuous depth representation \cite{wang2023learning}. Furthermore, Recurrent Structure Attention Guidance has been proposed \cite{yuan2023recurrent} along with the introduction of deep anisotropic diffusion for guided depth super-resolution \cite{metzger2023guided}. In addition, SGNet, a method that leverages gradient-frequency awareness, was created \cite{wang2024sgnet}.

Edge-guided studies \cite{xie2015edge, ye2018depth, liu2020edge, zhou2018depth} have played a pivotal role in advancing depth map super-resolution. However, these methods encounter challenges when transitioning to point clouds owing to varying settings and requirements. Inspired by the successes and limitations of these predecessors, our method embraces the concept of RGB guidance but redirects its focus directly towards point clouds. By doing so, we aim to achieve more precise and accurate upsampling results, bridging the gap between RGB-guided tasks and point cloud super-resolution.

\subsection{RGB-D and Point Cloud Datasets} Numerous datasets are commonly employed for GDSR and point cloud upsampling tasks. For GDSR, ScanNet \cite{dai2017scannet} stands out, offering 2.5 million RGB-D views derived from 1,513 indoor scans, capturing real-world environments using RGB-D cameras. Another notable dataset is the NYU Depth Dataset V2 \cite{silberman2012indoor}, which encompasses RGB-D images from various indoor scenes captured by depth cameras. The RGB-D-D dataset \cite{he2021towards}  comprises real-world paired low-resolution (LR) and super-resolution (HR) depth maps. However, a limitation of these extensive RGB-D datasets is that they provide only depth maps, necessitating conversion to point clouds for direct operation, thus making them less suitable for point-cloud-specific tasks.

In the realm of point cloud upsampling, PU-GAN \cite{li2019pu} and PU1K \cite{qian2021pu} are the most prevalent datasets. These are primarily synthetic and may not accurately represent the point cloud characteristics captured by RGB-D cameras in complex environments. Often, these synthetic datasets overlook practical challenges, such as noise, stray light effects, and data incompleteness. To bridge this gap, we curated a dataset using RGB-D cameras, focusing on a selected number of objects specifically tailored for training purposes. This dataset addresses a crucial requirement in the field by providing realistic point cloud data captured in complex environments.

\begin{figure*}[htpb]
\centering
\includegraphics[width=1.0\textwidth]{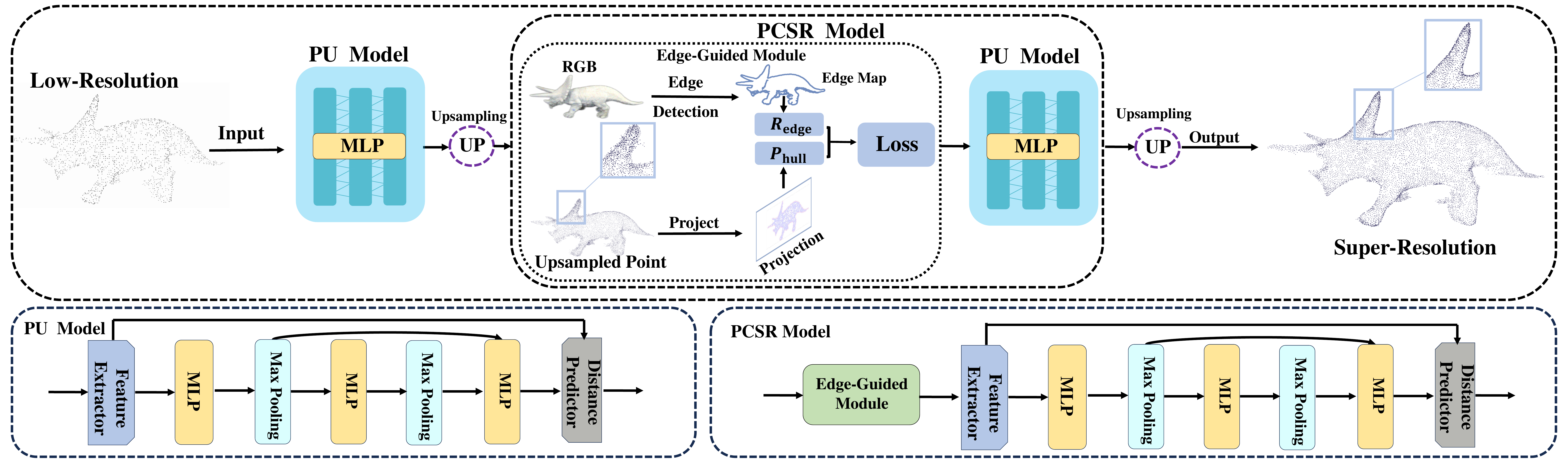}
\caption{The pipeline of our method starts with a PU model \cite{li2024learning} to increase point density, followed by the PCSR model for edge geometric optimization. The core innovation, the Edge-Guided Module, enhances point cloud boundaries using RGB edge information and refines the super-resolution process. The figure’s lower part illustrates how cascaded MLP blocks in both models progressively refine the point cloud.}
\label{fig:overview}
\end{figure*}

\section{Methods}

\subsection{Overview}
EGP3D was designed to process the initial low-resolution point cloud $P_{\text{low}}$ and transform it into a super-resolution point cloud $P_{\text{sup}}$. An overview of this method is shown in Figure \ref{fig:overview}. This section begins by detailing the architecture of the PU model, which increases the density of the sparse point cloud via upsampling. Next, we explain the process of mapping the upsampled point cloud onto the coordinate system of the RGB image, facilitating the extraction of its concave hull and alignment with the edge map of the RGB image. This integrated approach termed the edge-guided module, transforms the task into a 2D geometric optimization problem. To solve this problem, we introduce three loss functions: Chamfer distance loss, Hausdorff distance loss, and gradient smooth loss. Finally, we discuss the implementation of these loss functions within the point cloud upsampling model and overall loss functions.

\begin{figure*}[!t]
\centering
\includegraphics[width=1.0\textwidth]{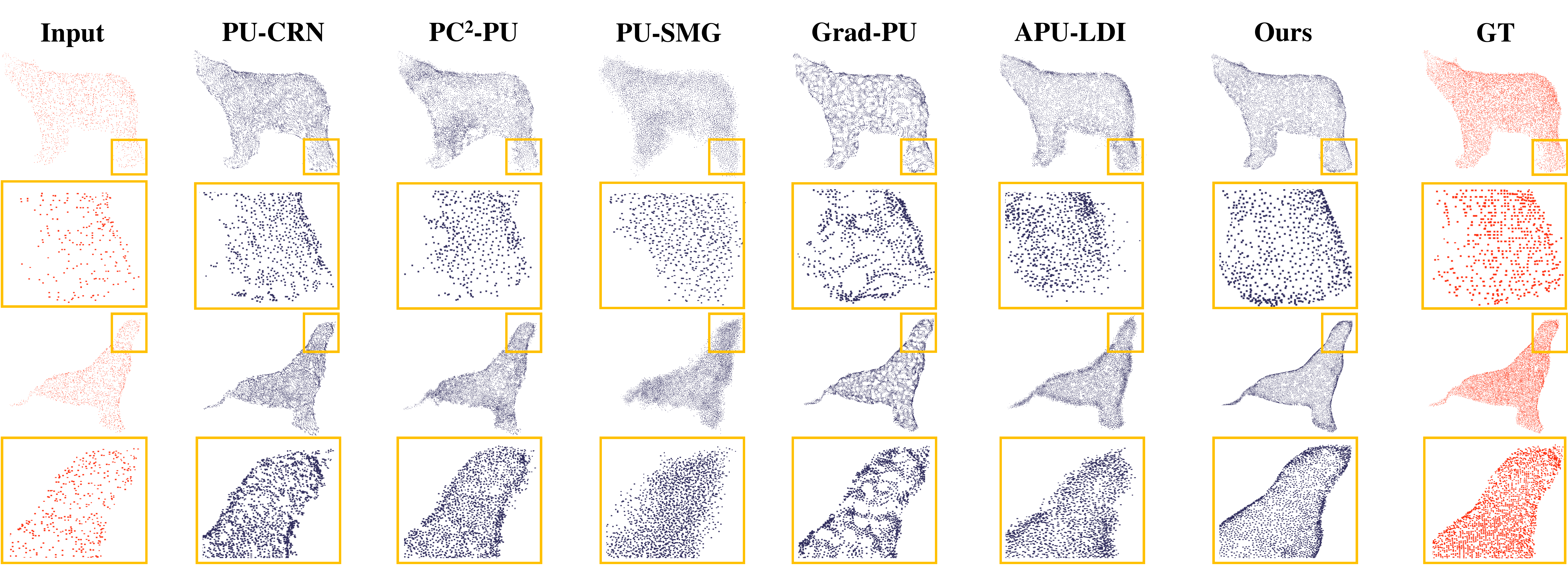} 
\caption{Qualitative comparisons of 4$\times$ upsampling results on the EGP3D dataset. Our method generates point clouds with clearer and more complete boundaries than state-of-the-art methods.}
\label{fig:experiment_compared}
\vspace*{-10pt}
\end{figure*}

\subsection{PU Model}
To achieve point cloud super-resolution, the first step is to increase the density of the low-resolution point cloud using upsampling. Below, we describe the network architecture of the PU model and the point cloud upsampling process.

\subsubsection{Network Architecture.} We implemented our method using APU-LDI \cite{li2024learning}. The network, known as the Local Distance Indicator (LDI), is shown in Figure \ref{fig:overview}. First, the sparse point cloud $S$ is divided into multiple local patches $P$. For each patch $P$, query points $q$ are generated around each point following a normal distribution. These query points, along with their corresponding local patches $P$, are then input into the network.  

Local patch $P = \{p_h\}_{h=1}^n$ is processed using a feature extractor to obtain point-wise features $F = \{f_h\}_{h=1}^n$. Using the $k$-nearest neighbor (KNN) algorithm \cite{he2023grad}, a smaller region $P_q = \{p_l\}_{l=1}^k$ is identified around query point $q$, and the points in $P_q$ are centralized to query point $q$. Then, the relative features $F' = \{f'_d\}_{d=1}^k$ were extracted using a multilayer perceptron (MLP).  

Next, the relative weights $w_d$ between the query point and its neighboring points are predicted using an attention mechanism. The final feature $f_q$ is obtained by weighting the relative and local features, as follows: 

\begin{equation}
f_q = \sum_{d=1}^{k} w_d \cdot f'_d + (1 - w_d) \cdot f_d
\end{equation}

Finally, the feature $f_q$, global feature of the local patch, and query point coordinates are concatenated and input into a distance predictor. MLP maps the final feature vector to the point-to-surface distance $d_l$. 

\subsubsection{Point Cloud Upsampling.} With the learned LDI, where the local distance $d_l$ provides crucial geometric details, the model learns the global implicit field. This field was constructed by generating a set of query points through uniformly sampled offsets added to each point in $P_L$. These query points are projected onto the implicit surface using the trained neural network $g_{\phi}$, which predicts both the distance and the gradient to the surface. The points are iteratively moved along the gradient direction until they align precisely with the surface, resulting in an upsampled point cloud, $P_U$.

\subsection{Edge-Guided Module}
After obtaining the upsampled point cloud $P_U$, it is necessary to perform edge-guided geometric optimization on its boundaries. First, $P_U$ must be aligned with the RGB image in a unified 2D coordinate system. This section introduces the alignment process, which functions as an edge-guided module integrated into subsequent computational modules. The details of this section's calculation  can be found in the Appendix.

\subsubsection{Point Cloud Projection.}
To leverage the edge information from the RGB image, we initially extracted the edge map, denoted by $R_{\text{edge}} = \{(x_i^{\text{edge}}, y_i^{\text{edge}})\}_{i=1}^{N_1}$, using established edge detection techniques \cite{rong2014improved, xie2015holistically, poma2020dense}. To use the RGB image captured by the RGB-D camera for guidance, we mapped the point cloud to the coordinate system of the RGB image. Given that our RGB-D camera employs the TOF scheme \cite{foix2011lock} to capture point clouds, we used the intrinsic and extrinsic parameters of the camera for this projection. Specifically, the upsampled point cloud $P_U = \{l_i\}_{i=1}^{N_2} = \{(x_i, y_i, z_i)\}_{i=1}^{N_2}$ is transformed into the point set $P_{\text{image}} = \{(u_i, v_i)\}_{i=1}^{N_3}$ on the RGB image plane.

\subsubsection{Concave Hull Computation.}
Next, the edge map of the RGB image was integrated with the projected point set $P_{\text{image}}$. The concave hull method \cite{park2012new} is used to compute the edges of this point set, providing a boundary that closely follows the shape of the point set. The K-nearest neighbors approach \cite{moreira2007concave} was employed to compute the concave hull, yielding the final boundary $P_{\text{hull}} = \{(x_i^{\text{hull}}, y_i^{\text{hull}})\}_{i=1}^{N}$.

\subsubsection{Unified Coordinate Framework.}
To ensure computational accuracy, the pre-processed data must be aligned within a unified coordinate system. This involves positioning both $R_{\text{edge}} = \{(x_i^{\text{edge}}, y_i^{\text{edge}})\}$ and $P_{\text{hull}} = \{(x_i^{\text{hull}}, y_i^{\text{hull}})\}$ within a consistent framework, as shown in the figure. By achieving this alignment, the original 3D point cloud optimization problem is transformed into an edge-guided optimization problem involving two sets of 2D points:
\begin{equation}
\min_{P_{\text{hull}}} \left\{ f(R_{\text{edge}}, P_{\text{hull}}) \right\},
\label{eq:problem1}
\end{equation}
where $R_{\text{edge}}$ represents the edge points extracted from the RGB image, $P_{\text{hull}}$ represents the points constituting the concave hull of the point cloud, and $f(R_{\text{edge}}, P_{\text{hull}})$ is a similarity measure between the two point sets.

\begin{figure*}[!t]
\centering
\includegraphics[width=1.0\textwidth]{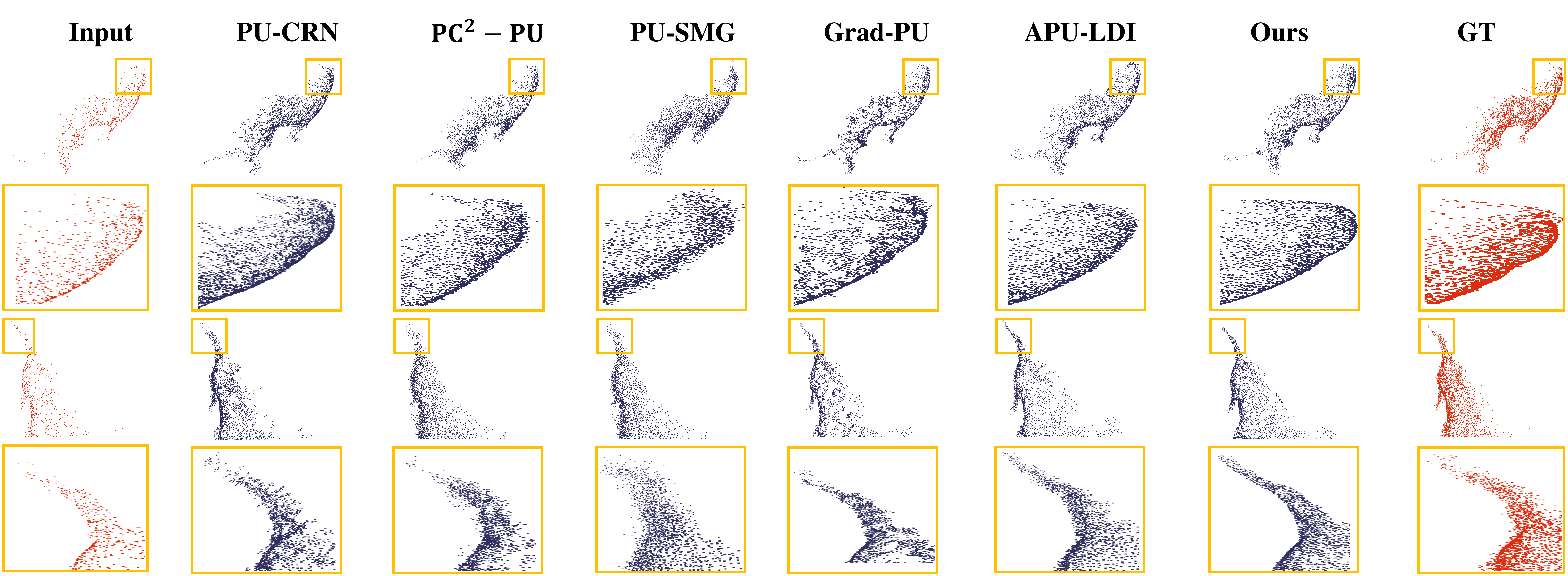}
\caption{The rotated point cloud visualization. From different viewpoints, it can still be observed that the point cloud generated by our method outperforms others in terms of resolution and the continuity of boundaries.}
\label{fig:dif_angle}
\vspace*{-10pt}
\end{figure*}

\subsection{Geometric Optimization}
To address Eq. \eqref{eq:problem1}, we designed three loss functions that target the shapes of 2D point set edges: Chamfer distance loss \cite{remy2002medial}, Hausdorff distance loss \cite{rucklidge1996efficient}, and Gradient Smooth Loss. The purpose and formulation of these loss functions are outlined below.

\subsubsection{Chamfer Distance Loss.}
The Chamfer distance loss quantifies the similarity between two point sets by summing the distances from each point in one set to its nearest point in the other. This metric is particularly effective for comparing the geometric structures of different point sets. In Eq. \eqref{eq:problem1}, we calculate the Chamfer distance loss between the edge points of the RGB image $R_{\text{edge}}$ and the hull points of the point cloud $P_{\text{hull}}$. The Chamfer distance between two sets, $R_{\text{edge}} = \{(x_i^{\text{edge}}, y_i^{\text{edge}})\}$ and $P_{\text{hull}} = \{(x_i^{\text{hull}}, y_i^{\text{hull}})\}$, is expressed as:
\begin{equation}
L_{\text{CD}}(R, P) = \sum_{a \in R} \min_{b \in P} \|a - b\|^2 
+ \sum_{b \in P} \min_{a \in R} \|b - a\|^2.
\end{equation}
This loss function ensures that the upsampled point cloud aligns closely with the edge information extracted from the RGB image, enhancing the fidelity and precision of the super-resolution process.

\subsubsection{Hausdorff Distance Loss.}
Hausdorff Distance Loss measures the maximum distance from a point in one set to its nearest neighbor in the other, capturing the worst-case scenario in aligning two geometric structures. Unlike Chamfer Distance, which averages all point-to-point distances, Hausdorff Distance emphasizes the most significant discrepancy between the sets.

In our framework, Hausdorff Distance Loss is defined as:
\begin{equation}
L_{\text{HD}}(R, P) = \max\left( d(R, P), d(P, R) \right),
\end{equation}
\begin{equation}
d(R, P) = \max_{{a \in R}{}} \min_{{b \in P}{}} \|a - b\|,
\end{equation}
\begin{equation}
d(P, R) = \max_{{b \in P}{}} \min_{{a \in R}{}} \|b - a\|.
\end{equation}
This loss function reduces the maximum deviation between the upsampled point cloud and the edge information from the RGB image, improving the alignment and accuracy of the super-resolution process.

\subsubsection{Gradient Smooth Loss.}
While the previous loss functions focus on optimizing shape similarity between point sets, Gradient Smooth Loss is designed to ensure smoothness of the point cloud boundaries during the super-resolution process. The mathematical formulation is:
\begin{equation}
L_{\text{GS}}(P_{\text{hull}}) = \sum_{i=1}^{N-2} \|\Delta g_i\| = \sum_{i=1}^{N-2} \|g_{i+1} - g_i\|,
\end{equation}
where \(|\Delta g_i|\) denotes the gradient change.
This function minimizes abrupt changes in gradients, promoting smoother transitions between points and enhancing the overall boundary the smoothness of the upsampled point cloud.

\subsection{PCSR Model}
After computing the loss, we implemented our PCSR model. The PCSR model was designed to address the limitations of the PU model in edge geometric optimization, as illustrated in Figure \ref{fig:overview}. This approach involves inputting the point cloud upsampled by the PU model into the edge-guided module to compute the designed loss, which is then used to update the PU model. This ensures that in subsequent upsampling, the PU model not only increases the point density but also optimizes the edges of the point cloud.
\subsubsection{Implementation Details.}
To fully realize our point cloud super-resolution model, it is essential to decouple the upsampling strategy from the model and integrate it, along with the three loss functions, into the training procedure. APU-LDI \cite{li2024learning} consists of two stages: the first stage involves training the LDI, whereas the second stage utilizes the LDI to learn the global implicit field. We integrated the edge-guided module into the second stage, using the upsampled point cloud $P_U$ as input. Edge optimization was then performed using the RGB edge map to update the global implicit field, resulting in a super-resolution point cloud with sharp edges.

\begin{table}[tp]
    \caption{Quantitative comparison between our method and the state-of-the-art point cloud upsampling methods with various scale factors on our EGP3D dataset.}
    \label{tab:PCD_exp} 
    \centering
    \begin{tabularx}{0.47\textwidth}{lXXXX}  
    \toprule 
    {Factor}&\multicolumn{2}{c}{$4\times$ (r=4)}   &\multicolumn{2}{c}{$16\times$ (r=16)}\cr
    \cmidrule{2-3} \cmidrule{4-5}
    \multirow{2}{*}{Methods} & CD$\downarrow$ & HD$\downarrow$ & CD$\downarrow$ & HD$\downarrow$ \cr
    & $10^{-1}$ & $10^{-2}$ & $10^{-1}$ & $10^{-2}$ \cr
    \midrule
    PU-CRN &0.183&0.467&0.423&0.529\cr
    PU-SMG &0.296&0.674&0.723&0.782 \cr
    {PC}$^2$-PU &0.201&0.355&0.392&0.664\cr
    Grad-PU &0.224&0.329&0.142&0.463\cr
    APU-LDI &0.172&0.289&0.126&0.314\cr
    \midrule
    Ours &\textbf{0.131}&\textbf{0.239}&\textbf{0.097}&\textbf{0.226} \cr
    \bottomrule
    \end{tabularx}    

\end{table}

\subsection{Loss Function}
The three designed loss functions are combined to form the overall loss function for our EGP3D model, expressed as:

\begin{equation}
L = \alpha L_{\text{CD}} + \beta L_{\text{HD}} + \gamma L_{\text{GS}},
\end{equation}
 where the coefficients $\alpha$, $\beta$, and $\gamma$ are set to $10^{-5}$, $10^{-2}$, and $10^{-2}$ respectively. These parameters are carefully adjusted to balance the contribution of each loss component to the total optimization process. In the corresponding upsampling models, the total loss is the sum of $L$ and the inherent loss function of the model.

\section{Experiments}
\subsection{Experiment Setup}
\paragraph{Datasets.} We captured geometry models from various angles using an Okulo P1 RGB-D camera, creating a dataset of 720 paired samples across 192 models. These models span a wide range of categories, including simple geometric objects, fruit models, and more complex forms.

In contrast to previous work \cite{he2023grad, li2024learning} that relied on synthesized data, our \textbf{EGP3D dataset} was captured under normal lighting conditions, making it more suitable for real-world RGB-D camera applications. The inherent noise, lighting effects, and varying model sensitivity to light eliminate the need for artificial data augmentation. Part of the data can be found in the appendix.

To evaluate our method, we conducted experiments using depth maps from the RGB-D-D dataset \cite{he2021towards}. We compared our approach with the previous GDSR \cite{metzger2023guided, wang2023learning} by converting the GDSR output depth maps into point clouds to maintain consistency in metric calculations.
\begin{figure*}[!t]
\centering
\includegraphics[width=1.0\textwidth]{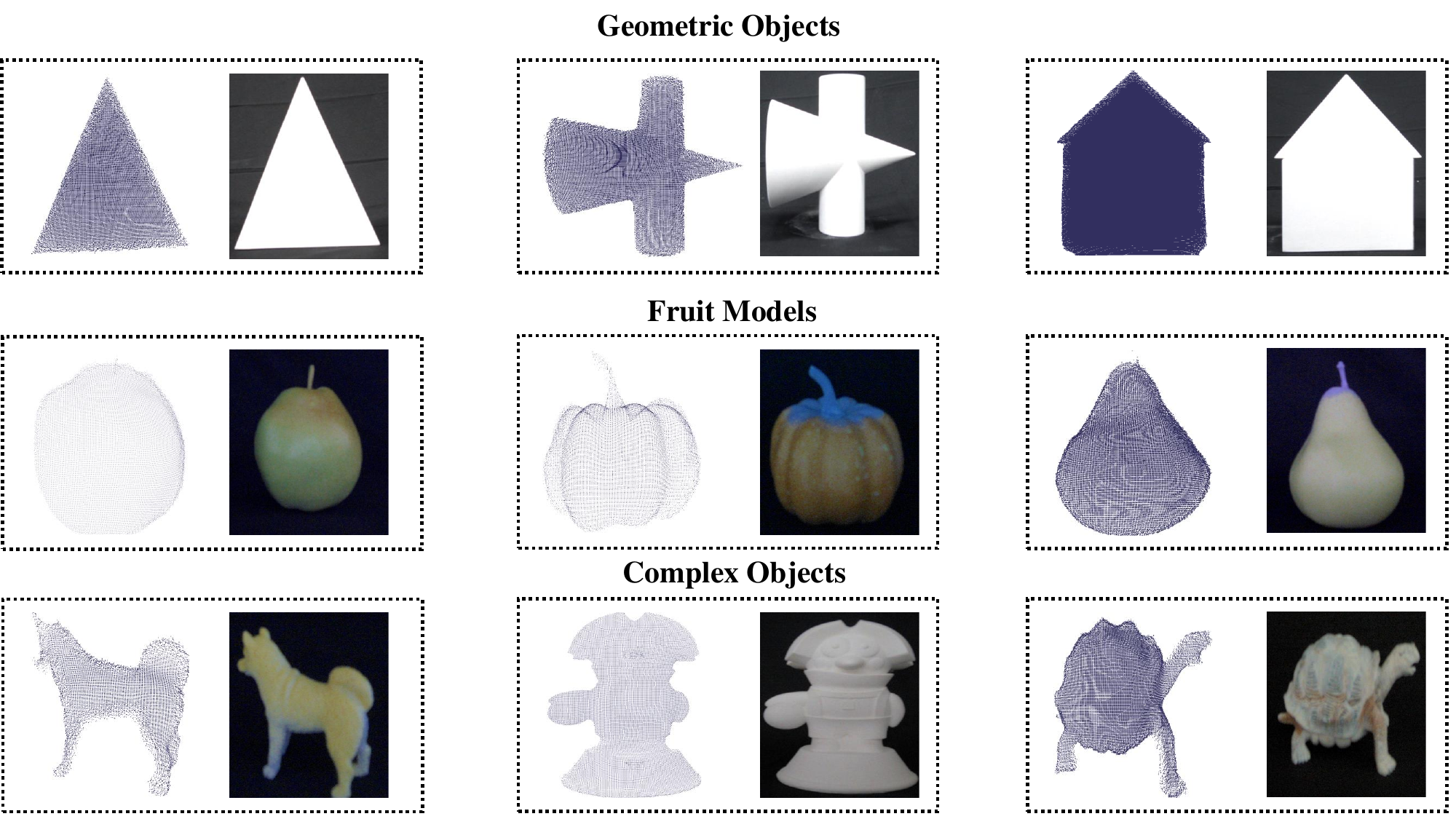}
\caption{A part of the EGP3D dataset showcasing point clouds and their corresponding RGB images, including: Geometric Objects, Fruit Models, and Complex Objects.
}
\label{fig:dataset}
\end{figure*}

\paragraph{Evaluation Metrics.} To evaluate the proposed method, we utilized two commonly used metrics in point cloud benchmarks: the \textbf{Chamfer distance (CD)} \cite{li2021point} and the \textbf{Hausdorff distance (HD)} \cite{luo2021pu}. These metrics are directly calculated on the point cloud and provide a comprehensive assessment of the model's performance. 

\paragraph{Experiment Details.} For the EGP3D dataset, we used 568 data samples for training and reserved the remaining 152 samples for testing. Given the large number of points in the captured point clouds, we applied a binning strategy \cite{orts2013point} to downsample the data to the desired number of points. Specifically, for training, we employed pairs of patches, with sparse patches containing 256 points and dense patches containing 1,024 points. During testing, each input point cloud, generated from watertight meshes, contained 2,048 points, while the ground truth consisted of 8,096 points. For the RGB-D-D dataset, we used a pre-trained model based on depth maps for the GDSR method, while our model utilized the pre-trained model on EGP3D. We selected 30 depth maps for testing, where evaluation in our model involved converting the depth maps into point clouds before conducting the metrics. 

To ensure a fair comparison, we trained all models using the Adam optimizer \cite{kingma2014adam} for 60 epochs, with a minibatch size of 32 and a learning rate that linearly decayed from 0.001 to 0. All experiments were conducted on a single GeForce RTX 3090 GPU using the Pytorch framework.

\subsection{Evaluation}

To validate the effectiveness of our method, we conducted extensive comparisons with state-of-the-art techniques on both our captured EGP3D dataset and the publicly available RGB-D-D dataset.
For evaluation, we use APU-LDI\cite{li2024learning} as the baseline model and compare it with several leading methods, including PU-CRN \cite{du2022point}, PU-SMG \cite{dell2022arbitrary}, PC$^{2}$-PU \cite{long2021pc2}, Grad-PU \cite{he2023grad}, and APU-LDI \cite{li2024learning}. The experimental results are presented in Table \ref{tab:PCD_exp}.

Table \ref{tab:Depth_exp} presents the comparison in GDSR methods such as as SGNet \cite{wang2024sgnet}, DADA \cite{metzger2023guided}, and GeoDSR \cite{wang2023learning}, our method also achieves sota performance.

\begin{table}[!t]
    \caption{Quantitative comparison between our method and the state-of-the-art GDSR methods with various scale factors on the RGB-D-D dataset. The depth maps were converted into point clouds before calculating the metrics.}
    \centering
    \begin{tabularx}{0.47\textwidth}{lXXXX}  
    \toprule 
    {Factor}&\multicolumn{2}{c}{$4\times$ (r=4)}   &\multicolumn{2}{c}{$16\times$ (r=16)}\cr
    \cmidrule{2-3} \cmidrule{4-5}
    \multirow{2}{*}{Methods} & CD$\downarrow$ & HD$\downarrow$ & CD$\downarrow$ & HD$\downarrow$ \cr
    & $10^{-1}$ & $10^{-2}$ & $10^{-1}$ & $10^{-2}$ \cr
    \midrule
    DADA &5.14&2.58&4.79&2.39\\
    GeoDSR &4.96&2.43&4.13&2.17\\
    SGNet &4.89&3.01&3.39&2.63\\
    \midrule
    Ours &\textbf{4.23}&\textbf{2.30}&\textbf{3.36}&\textbf{2.01} \\
    \bottomrule
    \end{tabularx}    
    \label{tab:Depth_exp} 
\end{table}

\subsubsection{Point Cloud Upsampling.}
As shown in Table~\ref{tab:PCD_exp}, our method significantly outperforms the other comparison approaches in terms of quantitative metrics. To further substantiate these numerical findings, we conducted a comprehensive visual comparison, illustrated in Figures~\ref{fig:experiment_compared} and \ref{fig:dif_angle}, under various projection viewing angles. These additional perspectives help reveal the subtle structural differences and highlight the fidelity of the reconstructed geometries.

The visual results clearly demonstrate that other methods tend to produce point clouds with blurred, irregular, or discontinuous boundaries, ultimately leading to global inconsistencies and less visually coherent surfaces. In stark contrast, our proposed method generates point clouds characterized by increased continuity, more complete and densely distributed points, and boundary regions that closely approximate the geometric contours of the ground truth. This high level of shape fidelity and boundary sharpness underscores the effectiveness of our approach in delivering superior super-resolution point clouds.

\subsubsection{Guided Depth Map Super-Resolution.}
We further evaluate our method on the more challenging RGB-D-D dataset, as summarized in Table~\ref{tab:Depth_exp}. This dataset poses additional complexity and variability, making it an ideal testbed to measure the robustness and adaptability of our approach in diverse real-world scenarios.

The performance comparison underscores that performing the upsampling operation directly on point clouds yields consistently better results than first upsampling on depth maps and then converting them into point clouds. The latter approach inherently risks introducing inaccuracies during the depth-to-point transformation, often caused by imperfect filtering, thresholding, or interpolation processes that degrade spatial fidelity and clarity. In contrast, by operating directly on point clouds, our method starts from a more precise geometric foundation and avoids these conversion pitfalls. As a result, the final super-resolved point clouds are more reliable, structurally coherent, and visually consistent, demonstrating that direct point cloud processing is a more effective strategy for high-quality 3D data reconstruction.

\subsection{Ablation Study and Analysis}
\begin{figure}[!t]
\centering
\includegraphics[width=1.0\linewidth]{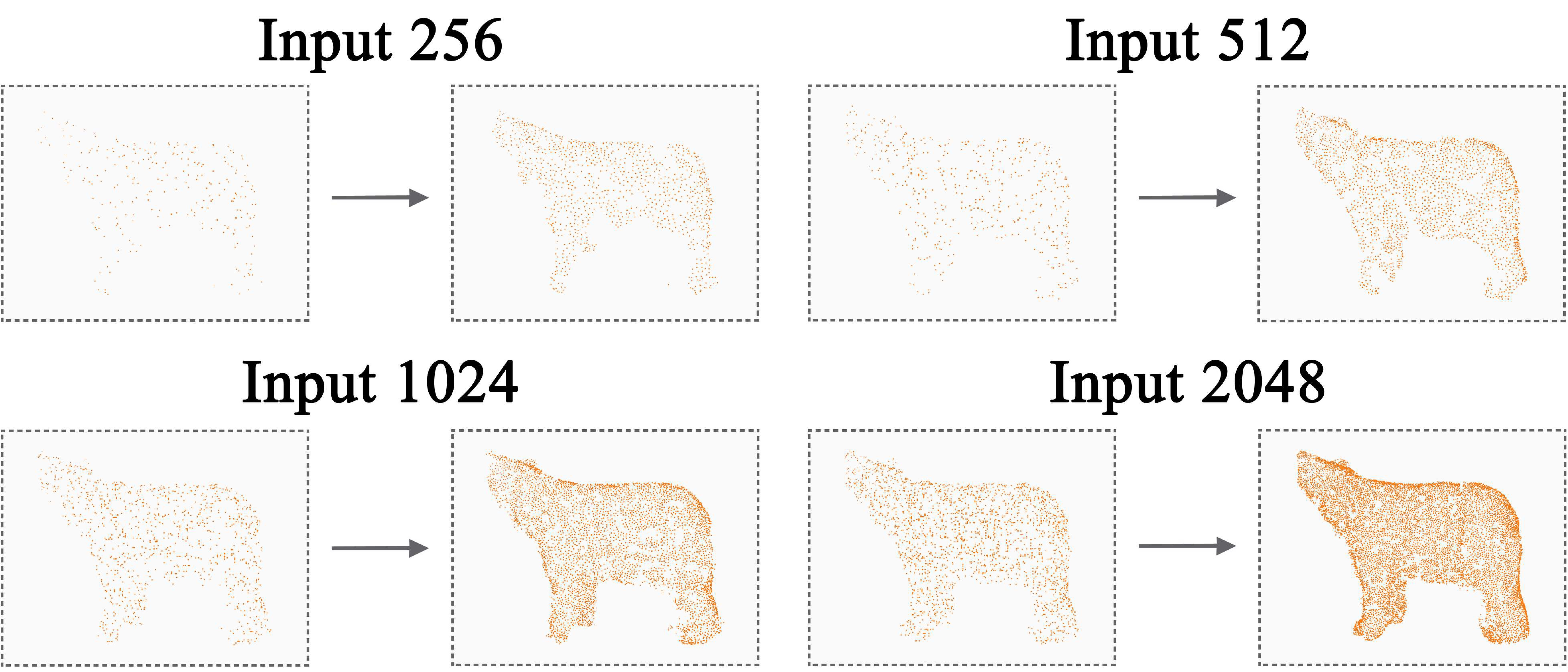}

\caption{4$\times$ results with various input sizes. The point cloud on the left is the input data, and the point cloud on the right is the point cloud after PCSR.}
\label{fig:various_input_size}
\end{figure}

\begin{table}[bp]
    \caption{Results of the ablation study for different experimental settings, including Chamfer, Hausdorff, and gradient smooth losses. Bold and underlined numbers indicate the best and second-best performance, respectively.}
    \centering
    \begin{tabularx}{0.47\textwidth}{XXXXX}
    \toprule
     $\mathcal{L}_{CD}$ & $\mathcal{L}_{HD}$ & $\mathcal{L}_{GS}$ & \multicolumn{1}{c}{\begin{tabular}[c]{@{}c@{}}CD $\downarrow$\\ $10^{-1}$\end{tabular}}    & \multicolumn{1}{c}{\begin{tabular}[c]{@{}c@{}}HD $\downarrow$\\ $10^{-2}$\end{tabular}} 
    \\
    \midrule
    $\checkmark$         &           &               & 0.147 & \underline{0.249} \\
              & $\checkmark$         &               & \underline{0.141} & 0.253 \\
              &           & $\checkmark$             & {0.185} & 0.262 \\
    $\checkmark$         & $\checkmark$         & $\checkmark$             & \textbf{0.131} & \textbf{0.239} \\
    \bottomrule
    \end{tabularx}
    \label{tab:Ablation_studies}
\end{table}

\subsubsection{Point Cloud Details Analysis.}
Our method is highly versatile and has been successfully applied in two different point cloud upsampling models \cite{he2023grad, li2024learning}. By examining Table \ref{tab:dif methods}, we can see that our method is capable of improving point cloud super-resolution performance based on its model. However, the degree of improvement remains constrained by the inherent quality of the original model’s performance. To ensure the fairness and generality of our experimental evaluations, we randomly select approximately 15\% of the data from the EGP3D dataset to serve as the test set for our experiments. As illustrated in Table~\ref{tab:Ablation_studies}, our approach enhances edge optimization in the super-resolution process across a range of models. It is worth noting that while Grad-PU \cite{he2023grad} tends to produce upsampled point clouds with certain holes and discontinuities, which somewhat limit the degree of boundary improvement, our method still provides a positive, albeit modest, enhancement. In contrast, for the more advanced APU-LDI \cite{li2024learning} model, which already generates high-quality point clouds, our method more effectively addresses boundary discontinuity issues, leading to more coherent and visually appealing edges.

Furthermore, our approach demonstrates robust performance even when dealing with sparse input point clouds, as evidenced by Figure~\ref{fig:various_input_size}. Regardless of the number of input points and their spatial distribution, our method consistently produces point clouds with sharp, rigid boundaries that remain unaffected by input sparsity. This adaptability underscores the reliability and practicality of our approach for diverse real-world scenarios, including those where the initial data might be incomplete or limited.

\begin{table}[!t]
    \caption{Quantitative comparison of applying EGP3D to Grad-PU \cite{he2023grad} and APU-LDI \cite{li2024learning}.}
    \centering
    \begin{tabularx}{0.47\textwidth}{lXXXX}  
    \toprule 
    {Factor}&\multicolumn{2}{c}{$4\times$ (r=4)}   &\multicolumn{2}{c}{$16\times$ (r=16)}\cr
    \cmidrule{2-3} \cmidrule{4-5}
    \multirow{2}{*}{Methods} & CD$\downarrow$ & HD$\downarrow$ & CD$\downarrow$ & HD$\downarrow$ \cr
    & $10^{-1}$ & $10^{-2}$ & $10^{-1}$ & $10^{-2}$ \cr
    \midrule
    Grad-PU &0.224&0.329&0.142&0.463\\
    APU-LDI &\underline{0.172}&0.289&\underline{0.126}&\underline{0.314}\\
    \midrule
    Grad-PCSR &0.195&\underline{0.277}&{0.136}&{0.401} \\
    APU-LDI-PCSR &\textbf{0.131}&\textbf{0.239}&\textbf{0.097}&\textbf{0.226} \\
    \bottomrule
    \end{tabularx}    
    \label{tab:dif methods} 
\end{table}

\subsubsection{Loss Function Analysis}
To thoroughly assess the contributions of our proposed loss function components, we conducted a series of ablation studies that examine various combinations of these loss terms. The results, summarized in Table~\ref{tab:Ablation_studies}, clearly indicate that employing the Chamfer distance loss and Hausdorff distance loss settings individually yields targeted improvements in the CD and HD metrics, respectively. Although the Gradient Smooth loss on its own does not significantly enhance the metrics, its true value emerges when combined with the other losses. In particular, the joint utilization of all three loss terms synergistically boosts the overall performance, achieving the highest level of accuracy and fidelity in the reconstructed point clouds.

\subsubsection{Real-World Visualization}
\begin{figure}[!t]
\centering
\includegraphics[width=1.0\linewidth]{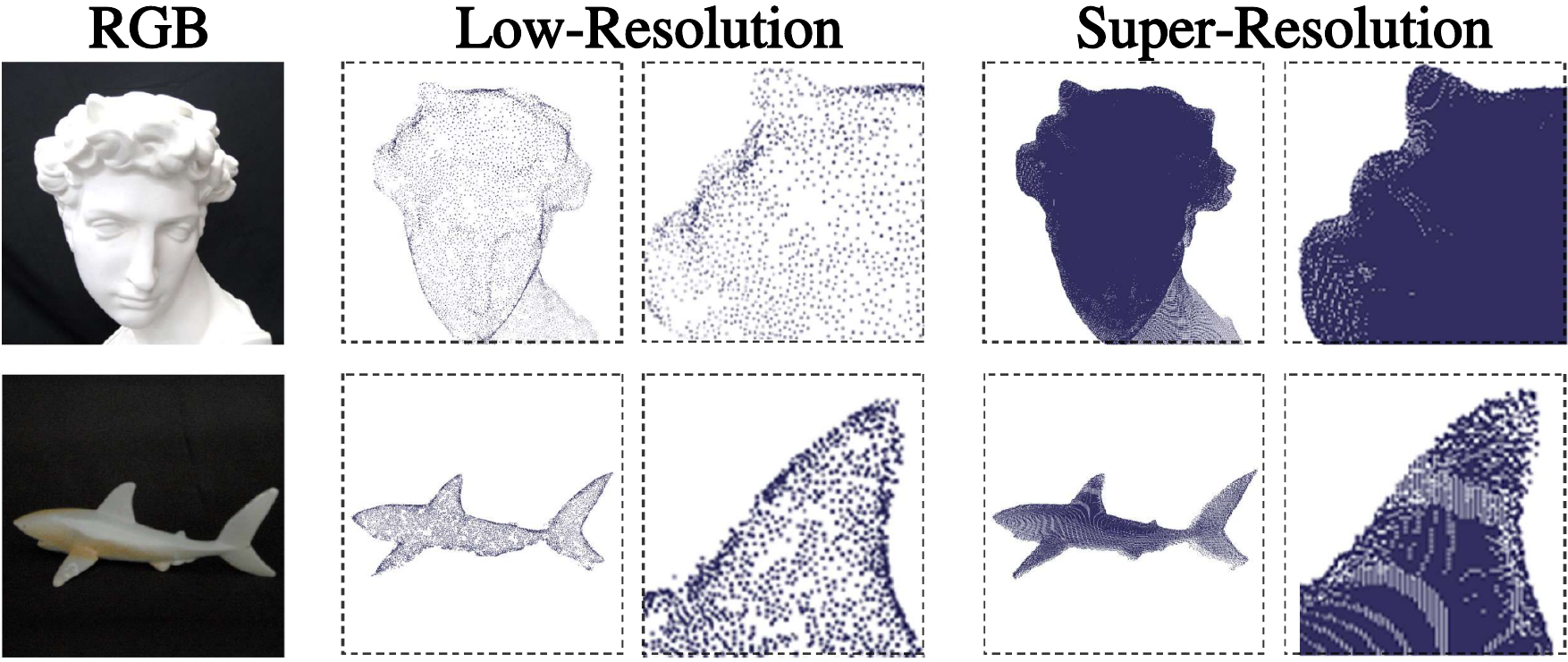}
\vspace{-0.5cm}
\caption{Super-resolution results with EGP3D method applied to low-resolution point clouds of real captures.}
\label{fig:real}
\vspace{-0.5cm}
\end{figure}

Our ultimate objective is to implement and validate the proposed method directly on practical hardware platforms, specifically RGB-D cameras, to ensure its applicability and effectiveness under real-world conditions. To demonstrate this, we applied our super-resolution technique to low-resolution point cloud data captured directly from an RGB-D camera without any intermediate conversions. The resulting point clouds, visualized in Figure~\ref{fig:real}, reveal that our method significantly enhances the density, continuity, and completeness of boundaries. In essence, the transformation from a low-resolution, potentially noisy input to a high-density, well-defined, and visually coherent super-resolution point cloud underscores the practical value and versatility of our approach.

\section{Limitation and Discussion}

While our work successfully introduces a novel point cloud super-resolution technique and the corresponding EGP3D dataset, it is important to acknowledge several limitations that currently restrain the method’s broader applicability and performance. One prominent issue arises from the characteristics of the data capture process: since RGB-D cameras typically record only one side of an object or scene, the resulting point clouds are often single-sided and incomplete. This limitation not only restricts the representational richness of the data but may also hinder downstream tasks such as holistic 3D reconstruction, where a complete, fully-enclosed representation of the object is critical.

Additionally, the need for precise camera calibration emerges as another challenge. Varying camera parameters, such as focal length, baseline, and exposure settings, can significantly influence the captured RGB-D data. Achieving accurate calibration becomes crucial when acquiring new datasets, as discrepancies between the RGB and depth channels can lead to misalignment, distortion, and a decline in overall point cloud fidelity. Furthermore, our method’s performance is inevitably constrained by the quality of the provided ground truth. When the reference data are imperfect, noisy, or incomplete, the super-resolution results may fail to achieve their full potential.

Looking ahead, future work could mitigate these challenges through multiple avenues. First, exploring multi-view capture techniques that synthesize information from multiple camera perspectives could offer a  more complete and balanced point clouds, thereby overcoming single-sided capture constraints. Second, the development of automated calibration procedures could streamline the data acquisition pipeline, enabling more reliable and user-friendly capture processes. Finally, further advancements in robust, adaptive algorithms that can gracefully handle variable ground truth quality would expand the practical utility of this approach, empowering it to generate high-fidelity results even in the face of imperfect reference data.

\section{Conclusion}
In this paper, we present an innovative edge-guided point cloud super-resolution technique, meticulously designed to craft high-resolution point clouds with exceptionally crisp and faithful boundary details. Our approach seamlessly integrates point cloud densification strategies with precise boundary refinement processes, ensuring that every edge is aligned closely and accurately with its true underlying form. By operating directly on point clouds, we circumvent the cumbersome and potentially error-prone depth-to-point conversion step, thus avoiding common pitfalls such as noise and artifacts that often arise from that process. This direct manipulation of point clouds represents a significant and pioneering step in the field, opening up new possibilities for enhanced clarity and structural fidelity in 3D representations.

Moreover, we have thoroughly validated the practicality of our method by applying it to real-world data captured with an RGB-D camera, demonstrating that our technique is not only theoretically sound but also robust and adaptable to real-life scenarios. Our extensive evaluations, conducted across a diverse array of datasets encompassing various object categories, scales, and complexity levels, consistently confirm the superior performance and efficacy of our proposed approach. As a result, our edge-guided point cloud super-resolution method establishes a new benchmark for quality and reliability in three-dimensional data processing, pushing the boundaries of what can be achieved in high-fidelity 3D reconstruction and analysis.

{\appendix
\section*{Edge-guided Module’s computational details}
In this section, we will introduce the computational methods used to unify the RGB edge map and point cloud projection in the same coordinate system within the edge-guided module. The section will cover the principles and algorithmic details, including edge detection, point cloud projection, and concave hull.
\subsection{Edge Detection}
In the selection of the edge detection method, after considering the complexity of computation and the effectiveness of the results, we chose the relatively simple yet effective Canny algorithm \cite{rong2014improved}.
The Canny edge detection algorithm consists of the following steps:

\begin{enumerate}
    \item \textbf{Gaussian Filtering:} Smooth the image with a Gaussian filter to reduce noise and unwanted details and textures. This is done by convolving the image with a Gaussian kernel:
    \begin{equation}
    I_{\text{smoothed}}(x, y) = I(x, y) * G(x, y, \sigma)
    \end{equation}
    where \( I(x, y) \) is the input image, \( G(x, y, \sigma) \) is the Gaussian kernel, and \( \sigma \) is the standard deviation of the Gaussian distribution.
    
    \item \textbf{Gradient Calculation:} Compute the gradient magnitude and direction at each pixel using finite differences:
    \begin{equation}
    G_x = \frac{\partial I}{\partial x}, \quad G_y = \frac{\partial I}{\partial y}
    \end{equation}
    \begin{equation}
    G = \sqrt{G_x^2 + G_y^2}, \quad \theta = \tan^{-1}\left(\frac{G_y}{G_x}\right)
    \end{equation}
    where \( G_x \) and \( G_y \) are the gradients in the \( x \) and \( y \) directions, \( G \) is the gradient magnitude, and \( \theta \) is the gradient direction.
    
    \item \textbf{Non-Maximum Suppression:} Thin the edges by keeping only local maxima in the gradient magnitude image, which means suppressing all other pixels that are not considered to be an edge.
    
    \item \textbf{Double Thresholding:} Apply a high and low threshold to identify strong, weak, and irrelevant edges. Pixels with gradient magnitudes above the high threshold are considered strong edges, while those below the low threshold are suppressed.
    
    \item \textbf{Edge Tracking by Hysteresis:} Finalize the edge detection by connecting weak edges to strong edges if they are in close proximity.
\end{enumerate}

\begin{figure}
\centering
\includegraphics[width=1\linewidth]{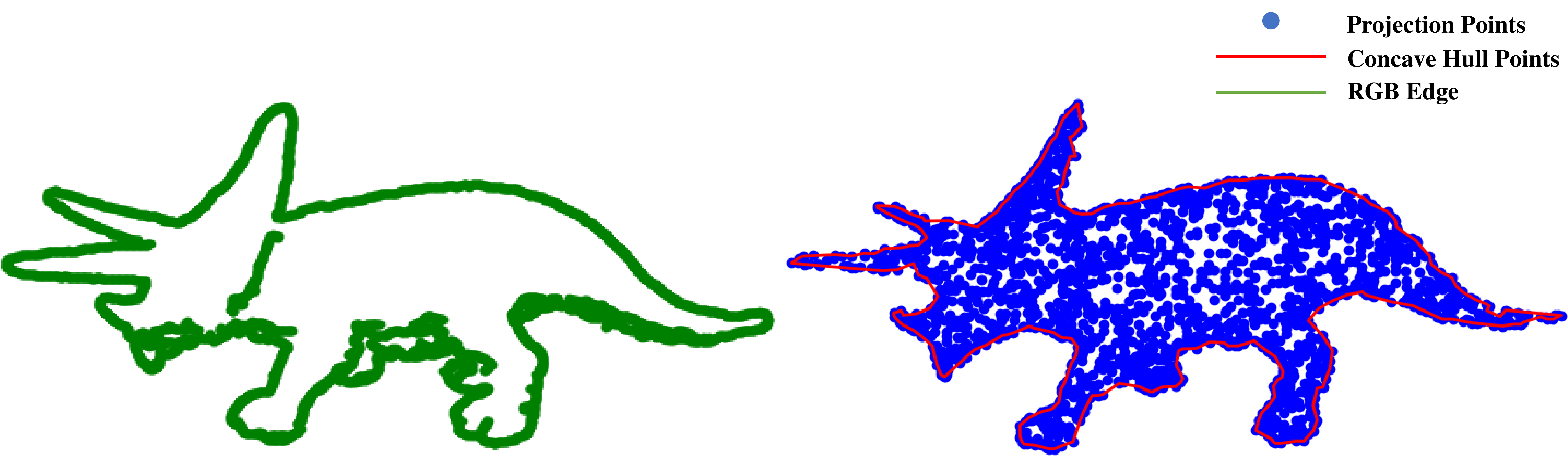}
\caption{The demonstration of point cloud projection and RGB edge map aligned in the same coordinate system, with the value of \(k\) set to 20 during the computation of the concave hull.}
\label{fig:edge_guided}
\end{figure}

\subsection{Point Cloud Projection}
To project the points from the point cloud onto the corresponding RGB image plane, we need to perform coordinate transformation and projection using the camera's intrinsic and extrinsic matrices. The following are the specific steps and their mathematical descriptions:

\begin{itemize}

\item \textbf{Load Intrinsic and Extrinsic Matrices}

The intrinsic matrix \(K\) defines the internal parameters of the camera, including focal length and the position of the principal point. We use the intrinsic matrix \(K_{\text{RGB}}\) of the RGB camera for this purpose.

The extrinsic matrix \(E\) defines the external parameters of the camera, including rotation and translation. For our process, we utilize the extrinsic matrix \(E_{\text{RGB}}\) for the RGB camera and the extrinsic matrix \(E_{\text{TOF}}\) for the point cloud camera.

\item \textbf{Coordinate Transformation}

First, we transform the point cloud data from the point cloud camera coordinate system to the RGB camera coordinate system. The point cloud data can be represented as \(P_{\text{TOF}} = \{(x, y, z)\}\), which needs to be expanded into homogeneous coordinates:

\begin{equation}
P_{\text{TOF}}^{\text{homo}} = \begin{pmatrix}
x \\
y \\
z \\
1
\end{pmatrix}
\end{equation}

Then, the point cloud coordinates are transformed into the RGB camera coordinate system using the extrinsic matrices:

\begin{equation}
P_{\text{RGB}}^{\text{homo}} = E_{\text{RGB}}^{-1} \cdot E_{\text{TOF}} \cdot P_{\text{TOF}}^{\text{homo}}
\end{equation}

\item \textbf{Projection onto the Image Plane}

The transformed point cloud coordinates \(P_{\text{RGB}}^{\text{homo}}\) are then projected onto the RGB image plane using the intrinsic matrix:

\begin{equation}
P_{\text{image}} = K_{\text{RGB}} \cdot \begin{pmatrix}
x_{\text{RGB}} \\
y_{\text{RGB}} \\
z_{\text{RGB}}
\end{pmatrix}
\end{equation}

To obtain the 2D coordinates on the image plane, normalization is performed:

\begin{equation}
u = \frac{x_{\text{image}}}{z_{\text{image}}}, \quad v = \frac{y_{\text{image}}}{z_{\text{image}}}
\end{equation}

where \( (u, v) \) are the pixel coordinates of the point cloud on the RGB image plane.

\item \textbf{Final Result}

Through the above steps, we map the original point cloud \(P_{\text{TOF}}\) to the point set \(P_{\text{image}}\) on the RGB image plane.

The entire process is summarized by the following mathematical formulas:

\begin{equation}
P_{\text{RGB}}^{\text{homo}} = E_{\text{RGB}}^{-1} \cdot E_{\text{TOF}} \cdot \begin{pmatrix}
x \\
y \\
z \\
1
\end{pmatrix}
\end{equation}

\begin{equation}
P_{\text{image}} = K_{\text{RGB}} \cdot P_{\text{RGB}}^{\text{homo}}
\end{equation}

\begin{equation}
(u, v) = \left( \frac{x_{\text{image}}}{z_{\text{image}}}, \frac{y_{\text{image}}}{z_{\text{image}}} \right)
\end{equation}

Thus, the mapping from the point cloud to the RGB image plane is completed.

\end{itemize}

\subsection{Concave Hull}
Our concave hull calculation method utilizes the K-Nearest Neighbors algorithm \cite{grapp07}. The main principles are as follows:

\begin{itemize}
    \item \textbf{Initialization:} Select an initial point \( p_0 \) from the set of points \( P_{\text{image}} \). This point serves as the starting point for constructing the concave hull.

    \item \textbf{K-Nearest Neighbors Search:} For each point currently in the hull \( P_{\text{hull}} \), find the \( k \)-nearest neighbors from the remaining points in \( P_{\text{image}} \). These neighbors are candidates for the next point to be added to the hull.

    \item \textbf{Selection of the Next Hull Point:} Among the \( k \)-nearest neighbors, select the point that, when added to the hull, does not create any line intersections with the existing edges of the hull. This ensures that the hull remains a simple polygon.

    \item \textbf{Iteration:} Repeat the process of selecting and adding points to the hull until all points in \( P_{\text{image}} \) have been considered. Remove each selected point from \( P_{\text{image}} \) as it is added to \( P_{\text{hull}} \).

    \item \textbf{Closing the Hull:} After all points have been processed, connect the last point in \( P_{\text{hull}} \) back to the initial point \( p_0 \) to close the hull and complete the concave boundary.
\end{itemize}

\begin{table*}[hb]
\caption{Comparison of the synthetic point cloud datasets PU-GAN \cite{li2019pu}, PU1K \cite{qian2021pu}, and EGP3D based on various criteria.}
\label{tab:comparison}
\centering
\resizebox{\textwidth}{!}{%
\begin{tabular}{|l|l|c|c|c|c|c|c|}
\hline
\textbf{Name} & \textbf{Method} & \textbf{Categories} & \textbf{Number} & \textbf{Single Objects} & \textbf{Single Side} & \textbf{Multi View} & \textbf{Generalizability} \\ \hline
PU-GAN & Synthesized & 147 & 147 & Yes & No & No & Low \\ \hline
PU1K & Synthesized & 1147 & 1147 & Yes & No & No & Medium \\ \hline
EGP3D & RGB-D & 192 & 720 & Yes & Yes & Yes & High \\ \hline
\end{tabular}%
}
\end{table*}

\section*{EGP3D Dataset}
Synthetic point cloud datasets for point cloud upsampling (PU) tasks overlook practical challenges such as noise, stray light effects, and data incompleteness. To bridge this gap, we curated a dataset named EGP3D using RGB-D cameras, focusing on a selected number of objects specifically tailored for training purposes.   

Our EGP3D method was compared with the commonly used PU task synthetic datasets PU-GAN \cite{li2019pu} and PU1K \cite{qian2021pu} from various aspects.  

The results can be seen in Table \ref{tab:comparison}. It can be observed from the table that the EGP3D dataset outperformed the PU-GAN dataset in various aspects. Although it has fewer categories and a smaller number of samples than PU1K, the EGP3D dataset, which was captured by an RGB-D camera and presented in a multi-view single-side format, demonstrates better practical generalizability. This makes it suitable for tasks related to RGB-D camera applications.  

We also present part of the EGP3D point-cloud dataset in Figure \ref{fig:dataset}. The point cloud in Figure \ref{fig:dataset} accurately reconstructs the shape and fine details of the RGB image.

\section*{Experimental Details}
In the comparative experiments, our method uses the following specific training parameters: a learning rate of 0.001, a maximum of 2000 iterations, warm-up ending at the 200th iteration, a batch size of 500, model saving frequency of 200 iterations, validation frequency of 50 iterations, and report frequency for each iteration, with \(\alpha\) set to 1.0, \(\beta_{\text{max}}\) set to 0.5, \(\gamma\) set to 0.1, and the random seed set to 2023. 

Our method was implemented on the global implicit field of the APU-LDI model \cite{li2024learning}, with the following parameter settings: output dimension of 1, input dimension of 3, hidden layer dimension of 512, eight network layers, skip connection at layer 4, multi resolution set to 0, bias set to 1.0, scale factor set to 1.0, geometric initialization set to true, and weight normalization set to true. 

}

\bibliography{ref}
\bibliographystyle{IEEEtran}

\end{document}